\documentclass[12pt]{article}

\RequirePackage{amsmath, amsthm, amssymb, mathtools, mathrsfs}
\RequirePackage{natbib}
\usepackage{dsfont} 
\usepackage{mathtools}
\mathtoolsset{showonlyrefs=true}
 
\allowdisplaybreaks

\title{Minimax Estimation of Discrete Distributions under Hockey-stick Divergence}

\usepackage[margin=1.2in]{geometry}
\usepackage{libertine}
\usepackage{courier}
\usepackage{mathpazo}
\usepackage{xcolor}
\usepackage{pifont}
\usepackage{authblk}
\usepackage[backref=page]{hyperref}
\usepackage{enumerate}

\usepackage{titlesec}

\definecolor{wred}{rgb}{0.533333,0.10980,0.15686}
\definecolor{wredlight}{rgb}{0.788, 0.11, 0.157}

\definecolor{darkbrown}{RGB}{170,100,0} 
\definecolor{darkdarkbrown}{RGB}{110,70,0} 
\titleformat{\section}
{\color{wred}\normalfont\Large\bfseries}
{\color{wred}\thesection}{1em}{}
\titleformat{\subsection}
{\color{wred}\normalfont\Large\bfseries}
{\color{wred}\thesubsection}{1em}{}

\hypersetup{
	colorlinks=true,
  linkcolor=wredlight,
  citecolor=wredlight,
  filecolor=wredlight,
  urlcolor=wredlight
}

 \renewcommand*{\backrefalt}[4]{
    \ifcase #1
     \or (page:~#2)
     \else (pages:~#2)
    \fi
    }

    \makeatletter
\def\@fnsymbol#1{\ensuremath{\ifcase#1\or \dagger\or \ddagger\or
   \mathsection\or \mathparagraph\or \|\or **\or \dagger\dagger
   \or \ddagger\ddagger \else\@ctrerr\fi}}
    \makeatother

\newtheorem{theorem}{Theorem}[section]
\newtheorem{lemma}{Lemma}[section]

\newtheorem{corollary}{Corollary}[section]
\newtheorem{remark}{Remark}[section]

\providecommand{\keywords}[1]
{
  \small	
  \textbf{\textit{Keywords---}} #1
} 

\DeclareMathOperator*{\argmin}{arg\,min}

\newcommand{\bbP}{\mathbb{P}}

\newcommand{\bbE}{\mathbb{E}}

\newcommand{\E}[2][]{\mathbb{E}_{#1}\left[ #2 \right]}

\newcommand{\Dir}{\mathrm{Dirichlet}}
\newcommand{\Betadist}[1]{\mathrm{Beta}\left( #1 \right)}
\newcommand{\Multinomial}{\mathrm{Multinomial}}
\newcommand{\Binomial}{\mathrm{Binomial}}
\newcommand{\TV}{\mathrm{TV}}
\newcommand{\Prob}[1]{\mathbb{P}\left(#1\right)}
\newcommand{\ind}[1]{\mathds{1}\left\{#1\right\}} 
\begin{document}

\author{Manaka Ueno\thanks{email: s234121x@st.go.tuat.ac.jp.} }

\author{Geoffrey Wolfer\thanks{email: wolfer@go.tuat.ac.jp.}}
\affil{Department of Electrical Engineering and Computer Science \\ Tokyo University of Agriculture and Technology, Tokyo}
\date{\today}

\maketitle

\begin{abstract}
We initiate the study of estimating a discrete distribution with respect to the hockey-stick divergence from independent samples. We provide non-asymptotic upper and lower bounds on the minimax expected loss in terms of the alphabet size, the number of samples, and the divergence parameter. The upper bounds are achieved by add-constant estimators, while the lower bound is derived via a Bayes risk argument. Both upper bounds and lower bounds  
reveal a transition between polynomial and exponential regimes as the parameters of the problem vary.
 \end{abstract}


\section{Introduction}
\label{section:introduction}
Estimating discrete probability distributions from samples is a fundamental task in statistics, information theory, and machine learning, having been the subject of extensive research \citep{orlitsky2003always, paninski2004variational, han2015minimax, kontorovich2025distribution}.
Although the Law of Large Numbers guarantees that we could 
reach a good estimate after infinite samples, 
in reality, this is impossible; 
we can only collect a finite number of samples. 
This raises the following question: 
how much data do we actually need to estimate a distribution well? 
One way to formalize this question 
consists in introducing a loss 
and analyzing its behavior 
in the minimax framework. 

While Total Variation distance is a well-studied  metric for distribution estimation, modern applications often require stronger or parameterized loss functions. A key example is the hockey-stick divergence (henceforth $E_\gamma$ divergence; HS). 
HS quantifies the probability mass where the likelihood ratio exceeds the threshold $\gamma$. 
HS plays a central role in various domains, including differential privacy \citep{dwork2006calibrating, kairouz2015composition}, hypothesis testing, and finite blocklength information theory \citep{polyanskiy2010channel}.

Although HS belongs to the family of $f$-divergences \citep{Csiszar1967} with generator 
$f(t) = (t - \gamma)_+$, existing minimax results 
for $f$-divergences \citep{kamath2015learning} fail 
to treat HS with $\gamma > 1$.  
Their analysis is  critically based on strict local convexity $f''(1) > 0$, whereas the HS generator is locally flat near $t=1$ ($f''(1) = 0$). Consequently, determining the minimax risk under HS cannot be deduced from prior theory and remains an open problem. 

In this work, we resolve this open problem by establishing upper and lower bounds on the minimax risk for discrete distribution estimation under the hockey-stick divergence.

\section{Notation and background}
\label{section:background}
Let $\Delta_k$ denote the set of discrete probability distributions over 
$[k] \coloneqq \{1, 2, \dots, k \}$, 
that is, the $(k-1)$-dimensional probability simplex defined as
    \begin{equation}
        \Delta_k \coloneqq
        \left\{
            p \in \mathbb{R}^{k}_{\ge 0}: 
            \sum_{i = 1}^{k} p_i = 1
        \right\} . 
    \end{equation}
Consider the problem of estimating an unknown distribution 
$p = (p_1, \dots, p_k) \in \Delta_k$ 
based on $n$ independent and identically distributed (i.i.d.) samples $X^n = (X_1, \dots, X_n)$ drawn from $p$. 
We construct an estimator $q = q(X^n) \in \Delta_k$ based on these samples.
Given a loss $L$, our interest is the worst expected loss achieved by the best estimator, 
called the minimax risk, 
    \begin{equation}
        R^L (k, n) \coloneqq 
        \underset{q}{\min} \, \underset{p \in \Delta_k}{\max}
        \, \underset{X^n \sim p}{\bbE}
        [L(p , q(X^n))] . 
    \end{equation}

In this paper, we focus on the minimax risk under the hockey-stick divergence 
($E_\gamma$ divergence; HS) , 
which is defined by 
    \begin{equation}
        E_\gamma(p \| q) \coloneqq 
        \sum_{i=1}^{k} (p_i - \gamma q_i)_+ ,
    \end{equation}
for $\gamma \ge 1$ and $(x)_+ = \max(x, 0)$ \citep{sason2016f}. 
Note that $E_\gamma$ achieves its maximum for $\gamma = 1$, where it coincides with the total variation distance (TV distance), defined as
    \begin{equation}
        \TV(p, q) \coloneqq 
        \frac{1}{2} \sum_{i=1}^k |p_i - q_i|. 
    \end{equation}
That is, we have
    \begin{equation}
        E_1(p \|q) = \TV(p, q) \geq E_\gamma(p \|q), \;\; \gamma \geq 1. \label{eq:HS-TV}
    \end{equation}

It is a classical result that the minimax risk for the TV distance decays at a rate of $O(\sqrt{k/n})$. 
More recently, \cite{kamath2015learning} %
settled the constants in the leading term 

    \begin{equation}
        R^{\TV}(k, n) = 
        \sqrt{\frac{(k-1)}{2 \pi n}} 
        + O\left( \frac{1}{n^{\frac{3}{4}}} \right) . 
    \end{equation} 
Given the equivalence in \eqref{eq:HS-TV}, 
it is natural to conjecture that as $\gamma$ approaches $1$, the minimax risk of HS will asymptotically exhibit a similar behavior. A complete survey of the problem of distribution estimation is beyond the scope of this paper, and we refer the reader to \citet[Section~1]{kamath2015learning} for a detailed account of the literature, including Kullback--Leibler and more general $f$-divergences, together with the historical developments.

\section{Main contributions}
\label{section:main-results}
In this section, 
we provide upper and lower bounds 
on the minimax risk under HS. 
 
\subsection{Upper Bounds}

We first present non-asymptotic upper bounds on the minimax risk under HS.

\begin{theorem} \label{theorem:upper}
    Let $n \ge 1$, $k \ge 1$, and $\gamma > 1$. 
    For $k \ge \gamma$, the minimax risk is bounded by
    \begin{align}
         R^{E_\gamma}(k, n) \, \le \, \min \left\{ \exp \left( - n \frac{(\gamma - 1)^2}{2 \gamma k} \right), R^{\TV}(k, n) \right\}. 
        \label{eq:upper:add-const}
    \end{align} 
    Conversely, for $k < \gamma$, the risk vanishes exactly:
    \begin{align}
         R^{E_\gamma}(k, n) = 0. 
    \end{align}
\end{theorem}

\begin{proof}[Proof sketch]
    Constructively, these two regimes are achieved by distinct estimation strategies. 
    The polynomial upper bound follows simply  from 
    the triangle inequality. 
    To get the exponential upper bound, we use 
    the Chernoff upper bound for a binomial random variable. 
Our proposed estimator is the add-$\lambda$ estimator $q_i = \frac{\frac{T_i}{n} + \frac{\lambda}{k}}{1+\lambda}, i \in [k]$, 
    with $\lambda = \frac{\gamma - 1}{2}$, 
    where $T_i$ is the count of samples of type $i$
    \[
        T_i \coloneqq \sum_{j=1}^n 
        \ind{X_j = i} . \label{eq:def-T_i}
    \]
The detailed proof is provided in Section~\ref{section:proof-upper-bound}.
\end{proof}

Theorem \ref{theorem:upper} 
shows a phase transition in the upper bound 
depending on the threshold parameter $\gamma$. 
When $\gamma$ is bounded away from $1$, an exponential rate of $O(\exp(-n/k))$ dominates. As $\gamma \to 1$, this bound weakens and transitions to the polynomial rate $O(\sqrt{k/n})$ governed by the TV distance.

\subsection{Lower Bounds}

Next, we provide the following non-asymptotic lower bounds on the minimax risk under HS.

\begin{theorem} \label{theorem:lower}
    Let $n \ge 1$, $1 < \gamma < k$.   
    Then, 
    \begin{align}
        R^{E_\gamma}(k, n) \ge \max \left\{ \gamma R^{TV}(k, n) - (\gamma - 1), \, \frac{k}{n + k}\left(
            1 - \frac{\gamma}{k}
        \right)^{n + k} \right\}.
        \label{eq:lower_bound}
    \end{align} 
\end{theorem}

\begin{corollary} \label{corollary:lower}
    Let $n \ge 1$, $k \ge 2$, and $\gamma \in (1, 2)$.  
    Set $\phi(\gamma) = (1-\gamma/2)^2$. Then, 
    \begin{align}
        R^{E_\gamma}(k, n) \ge \max \left\{ \gamma R^{TV}(k, n) - (\gamma - 1), \, \phi(\gamma) \exp{\left(- \frac{n}{k} (1 - \log \phi(\gamma))\right)} \right\}.
        \label{eq:lower_bound}
    \end{align}
\end{corollary}

\begin{proof}[Proof sketch]
The first polynomial lower bound follows from 
an known inequality between HS and TV. 
The second exponential lower bound is proved using the Bayes-risk approach 
with uniform prior $\Dir(1, \dots, 1)$. 
We use the fact that the order statistics sampled from a uniform distribution
have marginals 
belonging to the Beta distribution family. We defer the detailed proof to Section~\ref{section:proof-lower-bound}.
\end{proof}

Similarly to the upper bounds, Theorem \ref{theorem:lower} exhibits two regimes depending on $\gamma$. 
As $\gamma \to 1$, the first terms provide a polynomial lower bound of $\Omega(\sqrt{k/n})$, matching the rate of the upper bound. When $\gamma$ is larger, the second term yields an exponential lower bound $\Omega(\exp(-n/k))$.

\begin{remark}
    \cite{kamath2015learning} established bounds on the minimax risk for general $f$-divergence using the Bayes-risk approach. 
    They showed that 
    the minimax risk under certain conditions is 
    $\frac{(k-1)f''(1)}{2n} + o\left(\frac{1}{n}\right)$ for the generator function $f$. 
    However, they use Taylor approximation to the $f$-divergence at 1, 
    which is not directly applicable to our problem. 
\end{remark}

\begin{remark}[Discussion on tightness]
    We observe that the lower bounds in Corollary~\ref{corollary:lower} match the upper bounds of Theorem~\ref{theorem:upper} in non-trivial regimes of parameters when $\gamma < 2$, and we leave the exact dependence in $\gamma$ for future work.
\end{remark}

\section{Proofs}
\label{section:proofs}

 In this section, we provide 
 the deferred proofs of our main results. 
 
\subsection{Proof of Theorem~\ref{theorem:upper}}
\label{section:proof-upper-bound}

In this subsection, we present the proof of Theorem~\ref{theorem:upper}. 

First, note that for $k \le \gamma$, the minimax risk is equal to zero 
by choosing the estimator to be the uniform distribution: 
$q_i = \frac{1}{k}, \, i \in [k]$. 
Thus, below, we focus on the case of $k >  \gamma$. 

The first polynomial upper bound is directly obtained by applying the triangle inequality, to obtain \eqref{eq:HS-TV}. 

Next, we prove the exponential upper bound. 
We will use the following lemma.

\begin{lemma}[{\citep[Theorem~4.5]{mitzenmacher2017probability}}]

\label{lemma:upper-chernoff-binomial}
    Let $T_i \sim \Binomial(n, p_i)$. 
    Set its expectation as $\mu_i = np_i$. 
    Then, 
    from the Chernoff upper bound 
    on the tail probability of Binomial random variable, 
    for $s_i < \mu_i$, 
    we have 
    \begin{align}
        \Prob{T_i \le s_i} \le \exp \left( 
        - \frac{(\mu_i - s_i)^2}{2 \mu_i}
        \right) . 
    \end{align}
\end{lemma}

Fix any $\lambda \in [0, \gamma -1]$ 
and consider an add-$\lambda$ estimator $q_i = \frac{\frac{T_i}{n} + \frac{\lambda}{k}}{1 + \lambda}, \, i \in [k]$. 
Then, the expected loss of the add-$\lambda$ estimator under HS is, 
    \begin{align}
       \sum_{i=1}^k \E{(p_i - \gamma q_i)_+} &= \sum_{i=1}^k \E{\left( p_i - \gamma \frac{\frac{T_i}{n} + \frac{\lambda}{k}}{1+\lambda} \right)_+} \\
        &= \sum_{i=1}^k \E{\left( p_i - \gamma \frac{\frac{T_i}{n} + \frac{\lambda}{k}}{1+\lambda} \right) 
        \ind{p_i - \gamma \frac{\frac{T_i}{n} + \frac{\lambda}{k}}{1+\lambda} \ge 0}} \\
        &\overset{(\star1)}{=} \sum_{i=1}^k \E{\left( p_i - \gamma \frac{\frac{T_i}{n} + \frac{\lambda}{k}}{1+\lambda} \right) 
        \ind{p_i - \gamma \frac{\frac{T_i}{n} + \frac{\lambda}{k}}{1+\lambda} \ge 0}} 
        \ind{p_i \ge \frac{\lambda}{\lambda + 1}\frac{\gamma}{k}}
        \\
        &\le \sum_{i=1}^k \E{p_i 
        \ind{p_i - \gamma \frac{\frac{T_i}{n} + \frac{\lambda}{k}}{1+\lambda} \ge 0}} 
        \ind{p_i \ge \frac{\lambda}{\lambda + 1} \frac{\gamma}{k}} \\
        &= \sum_{i=1}^k p_i \Prob{
        p_i - \gamma \frac{\frac{T_i}{n} + \frac{\lambda}{k}}{1+\lambda} \ge 0}
        \ind{p_i \ge \frac{\lambda}{\lambda + 1} \frac{\gamma}{k}} \\
        &\overset{(\star2)}{\le} \sum_{i=1}^k p_i \exp{\left(
        - \frac{n}{2p_i}
        \left(
        p_i \frac{\gamma - 1 - \lambda}{\gamma} + \frac{\lambda}{k}
        \right)^2
        \right)}
        \ind{p_i \ge \frac{\lambda}{\lambda + 1} \frac{\gamma}{k}} \\
        &\le \max_{p_i} \exp{\left(
        - \frac{n}{2p_i}
        \left(
        p_i \frac{\gamma - 1 - \lambda}{\gamma} + \frac{\lambda}{k}
        \right)^2
        \right)}
        \ind{p_i \ge \frac{\lambda}{\lambda + 1} \frac{\gamma}{k}
        } \\
        &= \exp \left( 
            - n \min_{p \in [c, 1]} f(p)
        \right), 
    \end{align}
where $f(p) = \frac{1}{2p} \left( p \frac{\gamma - 1 - \lambda}{\gamma} + \frac{\lambda}{k} \right)^2$, 
$c = \frac{\lambda}{\lambda+1}\frac{\gamma}{k}$. 
$(\star1)$ comes from the fact that $T_i \ge 0$ for all $i \in [k]$. 
$(\star2)$ follows from lemma~\ref{lemma:upper-chernoff-binomial}. 

Few calculations give $p^* =  \argmin_{p \in [c, 1]} f(p)= \frac{\gamma}{k}$ for $k \ge \gamma$.  

We optimize the upper bound by choosing $\lambda = \frac{\gamma-1}{2}$. 
Thus, for $k \ge \gamma$ we obtain 
    \begin{align}
        R^{E_\gamma}(k,n) 
        \le \exp \left(
            - \frac{n}{k} \frac{(\gamma-1)^2}{2\gamma }
        \right) .
    \end{align}  
\qed

\subsection{Proof of Theorem~\ref{theorem:lower}}
\label{section:proof-lower-bound}

Now, we provide the proof of  Theorem~\ref{theorem:lower}. 

The polynomial lower bound follows from 
the following lemma. 
\begin{lemma}[{\citep[Eq.~(28)]{asoodeh2021local}}]
    \label{lemma:HS-TV-ineq-in-LB} 
    For any distributions $p$ and $q$ on $[k]$, 
    and any $\gamma \ge 1$, 
    \begin{align}
        E_\gamma(p\|q) \ge 1 - \gamma (1 - \TV(p,q)) . 
    \end{align}
\end{lemma}

To get the exponential lower bound, we bound the minimax risk 
by the Bayes risk. 
Let a Bayesian prior $\pi$ on $P$ be $\Dir(1, \dots, 1)$, that is, a uniform distribution over 
distributions and 
$P \sim \pi$ and 
$X_i \sim P_i$. Recalling \eqref{eq:def-T_i}, this also means 
$T|P=(T_1, \dots, T_k)|P \sim \Multinomial(n, P)$.
The minimax risk can be lower bounded by the Bayes risk as follows: 
    \begin{align}
        R^{E_\gamma}(k,n)
        &= \underset{q}{\inf} \, \underset{p}{\sup} \,  \bbE_T [E_\gamma(p\|q(T))] \\
        &\ge \underset{q}{\inf} \,
             \bbE_{P} \bbE_{T} [E_\gamma(P\|q(T))] \\
        &= \underset{q}{\inf} \, \sum_{i=1}^k
            \bbE_{T} \bbE_{P|T} [(P_i - \gamma q_i)_+ |
            T]. 
    \end{align}
Fixing $T_1 = t_1, \dots, T_k=t_k$, 
we will find a lower bound for, 
    \begin{align}
        \bbE_{P|T} [(P_i - \gamma q_i)_+ |T_1 = t_1, \dots, T_k=t_k]. 
    \end{align} 
Note that the posterior is 
    \begin{align}
        P | (T_1=t_1, \dots, T_k=t_k) \sim 
        \Dir(T_1 + 1, \dots, T_k + 1),
    \end{align}
and in turn we obtain the marginal posterior distributions, 
    \begin{align}
        P_i|(T_1=t_1, \dots, T_k=t_k) 
        \sim \Betadist{t_i+1, n - t_i + (k-1)}. 
    \end{align}
We will use the following lemma that connects the Beta distribution to order statistics. 
 
\begin{lemma}[{\citep[Example~5.4.5]{casella2002statistical}}]
\label{lemma:beta-order-stat}
    Let $U_1, U_2, \dots, U_m \sim \mathrm{Uniform}([0,1])$ and 
    $1 \le r \le m$. 
    Define order statistics, 
        \[U_{(1)} \le U_{(2)} \le 
            \cdots \le U_{(m)} . \]
    Then, the $r$th order statistic satisfies
        \[
            U_{(r)} \sim \Betadist{r, m-r+1}. 
        \]
\end{lemma}

Below, we apply Lemma~\ref{lemma:beta-order-stat} with $r = T_i + 1$ and $m = n + k - 1$ 
at $(\star1)$. 
First, suppose that $\gamma q_i \le 1$. 
Then, 
    \begin{align}
        \bbE_{P|T}[(P_i - \gamma q_i(t_1, \dots, t_k))_+ | T_1=t_1, \dots, T_k=t_k]
        &= \int_{\gamma q_i}^{1} \Prob{P_i > u |T_1=t_1, \dots, T_k=t_k} du \\
        &\overset{(\star1)}{=} 
        \int_{\gamma q_i}^{1} \bbP_{U_1,\dots,U_{n+k-1}}\left(
            U_{(T_i+1)} > u
        \right) du \\
        &\ge \int_{\gamma q_i}^{1} 
        \bbP_{U_1,\dots,U_{n+k-1}}
        \left(
            U_{(1)} > u
        \right) du \\
        &= \int_{\gamma q_i}^{1} 
        \bbP_{U_1,\dots,U_{n+k-1}}
        \left(
            U_{1} > u, \dots, 
            U_{n + k - 1} > u 
        \right) du \\
        &= \int_{\gamma q_i}^{1} 
            (1-u)^{n+k-1} du \\
        &= \frac{(1 - \gamma q_i)^{n+k}}{n + k} .
    \end{align}
The crucial step stems from our choice of a prior and property of the Beta distribution. 
On the other hand, when $\gamma q_i > 1$, 
we have 
$\bbE_{P|T}[(P_i - \gamma q_i(t_1,\dots,t_k))_+|T_1=t_1, \dots, T_k=t_k] 
= 0$. 
Thus, combining the two cases, 
    \begin{align}
        \bbE_{P|T}[(P_i - \gamma q_i(t_1,\dots,t_k))_+
        |T_1=t_1, \dots, T_k=t_k] 
        \ge \frac{(1 - \gamma q_i)_{+}^{n+k}}{n + k} . 
    \end{align}
Then, summing over $i$, and with Jensen's inequality $(\star2)$, 
    \begin{align}
        \bbE_{P|T}[E_\gamma(P \| q(t_1,\dots,t_k))
        |T_1=t_1, \dots, T_k=t_k] 
        &\ge \frac{k}{n + k}
        \left(
            \frac{1}{k} \sum_{i=1}^{k} 
            (1 - \gamma q_i)_{+}^{n + k}
        \right) \\
        &\overset{(\star2)}{\ge} \frac{k}{n + k} \left(
        \left(
            1 - \gamma \frac{1}{k} \sum_{i=1}^k q_i
        \right)_{+}^{n + k}
        \right) \\
        &= \frac{k}{n + k} \left(
            1 - \frac{\gamma}{k}
        \right)^{n + k} \\
        &= \frac{1}{1 + r}\left(
            1 - \frac{\gamma}{k}
        \right)^{k (1 + r)}, \,\, r = n/k
    \end{align}
where we recall that $\gamma < k$ (otherwise the problem is trivial). 
The lower bound follows naturally. 
Regarding applying Jensen's inequality, 
define $f(u) = (1-\gamma u)_{+}^{n + k}$ and 
we use the fact that it is convex. 
Corollary~\ref{corollary:lower} is an application of the Bernoulli inequality for $1 < \gamma < 2, 
k \ge 2$, which gives $\left(1 - \frac{\gamma}{k} \right)^{k/2}
\ge 1 - \gamma/2$, and the inequality $(1+r)^{-1} \ge e^{-r}$.

\qed  
\bibliography{bibliography}

\begin{thebibliography}{13}
\providecommand{\natexlab}[1]{#1}
\providecommand{\url}[1]{\texttt{#1}}
\expandafter\ifx\csname urlstyle\endcsname\relax
  \providecommand{\doi}[1]{doi: #1}\else
  \providecommand{\doi}{doi: \begingroup \urlstyle{rm}\Url}\fi

\bibitem[Asoodeh et~al.(2021)Asoodeh, Aliakbarpour, and
  Calmon]{asoodeh2021local}
S.~Asoodeh, M.~Aliakbarpour, and F.~P. Calmon.
\newblock Local differential privacy is equivalent to contraction of
  {$E_\gamma$-Divergence}.
\newblock \emph{arXiv preprint arXiv:2102.01258}, 2021.

\bibitem[Casella and Berger(2002)]{casella2002statistical}
G.~Casella and R.~L. Berger.
\newblock \emph{Statistical inference}, volume~2.
\newblock Duxbury Pacific Grove, CA, 2002.

\bibitem[Csisz\'{a}r(1967)]{Csiszar1967}
I.~Csisz\'{a}r.
\newblock Information-type measures of difference of probability distributions
  and indirect observations.
\newblock \emph{Studia Scientiarum Mathematicarum Hungarica}, 2:\penalty0
  299--318, 1967.

\bibitem[Dwork et~al.(2006)Dwork, McSherry, Nissim, and
  Smith]{dwork2006calibrating}
C.~Dwork, F.~McSherry, K.~Nissim, and A.~Smith.
\newblock Calibrating noise to sensitivity in private data analysis.
\newblock In S.~Halevi and T.~Rabin, editors, \emph{Theory of Cryptography},
  pages 265--284, Berlin, Heidelberg, 2006. Springer Berlin Heidelberg.
\newblock ISBN 978-3-540-32732-5.

\bibitem[Han et~al.(2015)Han, Jiao, and Weissman]{han2015minimax}
Y.~Han, J.~Jiao, and T.~Weissman.
\newblock Minimax estimation of discrete distributions under $\ell_{1}$ loss.
\newblock \emph{IEEE Transactions on Information Theory}, 61\penalty0
  (11):\penalty0 6343--6354, 2015.
\newblock \doi{10.1109/TIT.2015.2478816}.

\bibitem[Kairouz et~al.(2015)Kairouz, Oh, and
  Viswanath]{kairouz2015composition}
P.~Kairouz, S.~Oh, and P.~Viswanath.
\newblock The composition theorem for differential privacy.
\newblock In F.~Bach and D.~Blei, editors, \emph{Proceedings of the 32nd
  International Conference on Machine Learning}, volume~37 of \emph{Proceedings
  of Machine Learning Research}, pages 1376--1385, Lille, France, 07--09 Jul
  2015. PMLR.

\bibitem[Kamath et~al.(2015)Kamath, Orlitsky, Pichapati, and
  Suresh]{kamath2015learning}
S.~Kamath, A.~Orlitsky, D.~Pichapati, and A.~T. Suresh.
\newblock On learning distributions from their samples.
\newblock In P.~Grünwald, E.~Hazan, and S.~Kale, editors, \emph{Proceedings of
  The 28th Conference on Learning Theory}, volume~40 of \emph{Proceedings of
  Machine Learning Research}, pages 1066--1100, Paris, France, 03--06 Jul 2015.
  PMLR.

\bibitem[Kontorovich and Painsky(2025)]{kontorovich2025distribution}
A.~Kontorovich and A.~Painsky.
\newblock Distribution estimation under the infinity norm.
\newblock \emph{Journal of Machine Learning Research}, 26\penalty0
  (162):\penalty0 1--30, 2025.

\bibitem[Mitzenmacher and Upfal(2017)]{mitzenmacher2017probability}
M.~Mitzenmacher and E.~Upfal.
\newblock \emph{Probability and Computing: Randomization and Probabilistic
  Techniques in Algorithms and Data Analysis}.
\newblock Probability and Computing: Randomization and Probabilistic Techniques
  in Algorithms and Data Analysis. Cambridge University Press, 2017.
\newblock ISBN 9781107154889.

\bibitem[Orlitsky et~al.(2003)Orlitsky, Santhanam, and
  Zhang]{orlitsky2003always}
A.~Orlitsky, N.~P. Santhanam, and J.~Zhang.
\newblock Always {G}ood {T}uring: Asymptotically optimal probability
  estimation.
\newblock \emph{Science}, 302\penalty0 (5644):\penalty0 427--431, 2003.
\newblock \doi{10.1126/science.1088284}.

\bibitem[Paninski(2004)]{paninski2004variational}
L.~Paninski.
\newblock Variational minimax estimation of discrete distributions under {KL}
  loss.
\newblock In L.~Saul, Y.~Weiss, and L.~Bottou, editors, \emph{Advances in
  Neural Information Processing Systems}, volume~17. MIT Press, 2004.

\bibitem[Polyanskiy et~al.(2010)Polyanskiy, Poor, and
  Verdu]{polyanskiy2010channel}
Y.~Polyanskiy, H.~V. Poor, and S.~Verdu.
\newblock Channel coding rate in the finite blocklength regime.
\newblock \emph{IEEE Transactions on Information Theory}, 56\penalty0
  (5):\penalty0 2307--2359, 2010.
\newblock \doi{10.1109/TIT.2010.2043769}.

\bibitem[Sason and Verdú(2016)]{sason2016f}
I.~Sason and S.~Verdú.
\newblock $f$-divergence inequalities.
\newblock \emph{IEEE Transactions on Information Theory}, 62\penalty0
  (11):\penalty0 5973--6006, 2016.
\newblock \doi{10.1109/TIT.2016.2603151}.

\end{thebibliography}
\bibliographystyle{abbrvnat}

\end{document}